\documentclass[runningheads]{llncs}

\usepackage{eccv}

\usepackage{eccvabbrv}

\usepackage{graphicx}
\usepackage{booktabs}

\usepackage[accsupp]{axessibility}  

\usepackage{booktabs}
\usepackage{amsmath}
\DeclareMathOperator*{\argmax}{arg\,max}

\usepackage{multirow}
\usepackage[table]{xcolor}
\usepackage{pifont}

\usepackage{amsmath}
\usepackage{amssymb}
\usepackage{algorithm}
\usepackage{algorithmicx}
\usepackage{algpseudocode}
\usepackage{wrapfig}
\usepackage{pdfpages} 

\newcommand{\cmark}{\ding{51}} 
\newcommand{\xmark}{\ding{55}} 

\newcommand{\tit}[1]{\smallbreak\noindent\textbf{#1.}}

\def \eg {\emph{e.g.}}
\def \etal {\emph{et al.}}

\newcommand{\ours}{{SLS}\xspace}

\definecolor{OurColor}{rgb}{0.855, 0.937, 0.957}

\newcommand{\smallhead}[1]{\textbf{\small #1}}

\definecolor{cvprblue}{rgb}{0.21,0.49,0.74}
\usepackage[pagebackref,breaklinks,colorlinks,allcolors=eccvblue]{hyperref}

\usepackage{hyperref}

\usepackage{orcidlink}

\begin{document}

\title{A Scalable Vector Graphics Latent Space}

\titlerunning{A Scalable Vector Graphics Latent Space}

\author{Leonardo Zini\inst{1}\orcidlink{0009-0003-9439-9867} \and
Elia Frigieri\inst{1} \and
Lorenzo Baraldi\inst{1}\orcidlink{0000-0001-5125-4957}}

\authorrunning{L.~Zini et al.}

\institute{University of Modena and Reggio Emilia, Italy \\
\email{\{name.surname\}@unimore.it} \\
\href{https://aimagelab.github.io/svg_latent_space/}{\url{aimagelab.github.io/svg_latent_space/}}
}

\maketitle

\begin{abstract}
Scalable Vector Graphics are a fundamental medium for resolution\nobreakdash-independent visual content, yet the deep learning community lacks a continuous, dense, and invertible latent space for vector representations, the kind of foundational building block that Variational Autoencoders and their descendants have long provided for raster images. We introduce \textbf{SLS} (\textbf{S}VG \textbf{L}atent \textbf{S}pace), a Transformer-based autoencoder that learns compact dense representations of individual SVG paths, the atomic visual elements from which any SVG image can be composed. By modeling SVG commands, coordinate data, and visual properties within a unified BPE-based token vocabulary, SLS learns fixed-size latent representations that jointly capture structure and appearance, and can be decoded back into valid, style-consistent SVG paths with high fidelity. The resulting embedding space is robust, invertible, and structured: embeddings lie on a unit hypersphere, enabling efficient similarity search, composition, and downstream conditioning through simple vector-space operations. Finally, we demonstrate that \ours generalizes across diverse tasks reducing their FLOPs by over $150\times$ compared to token-based approaches, and establishing a general-purpose latent foundation for vector graphics research. 
\end{abstract}

\section{Introduction}
\label{sec:intro}

Scalable Vector Graphics have become a foundational medium for resolution-independent visual content, spanning iconography, UI design, data visualization, and beyond. Unlike raster images, SVGs describe geometry symbolically through sequences of drawing commands, coordinates, and style attributes -- a representation that is inherently interpretable and editable, yet poses a fundamental challenge for deep learning: how do we build a continuous, compact, and semantically meaningful space over vector content?

This question has remained largely unanswered. Existing approaches either rely on rasterized renderings~\cite{hu2024supersvg, xing2024svgdreamer, jain2023vectorfusion}, indirectly learning representations in the pixel domain and thus discarding the underlying geometric structure, or process SVG markup as raw text sequences. The latter strategy, adopted by a growing family of LLM-based methods~\cite{we2023iconshop,zini2025vhector,xing2025empowering,yang2025omnisvg}, directly generates or consumes SVG code, but does not learn a continuous and compact latent space over vector content. Within these approaches, SVG elements are serialized into long token sequences, which leads to sparse, high-dimensional representations that strain context windows and complicate downstream tasks such as retrieval or captioning at scale. 
The reliance on rasterization or long token sequences is not incidental: it reflects the lack of a sufficiently expressive and scalable invertible embedding space tailored to vector content. Without a compact and semantically structured space, models must either discard symbolic structure by operating in the pixel domain or process SVG markup as sparse text sequences, both of which hinder scalability and downstream reasoning.
While the broader vision community has long benefited from latent embedding spaces for images, enabling retrieval, captioning, and generation through simple vector-space operations, an equivalent continuous and semantically structured embedding space for SVG content remains underexplored.
Existing approaches to SVG representation learning introduce latent spaces that are constrained in terms of embedding fidelity, input versatility, and stylistic coverage.

\begin{figure}[t]
    \centering
    \includegraphics[width=0.95\linewidth]{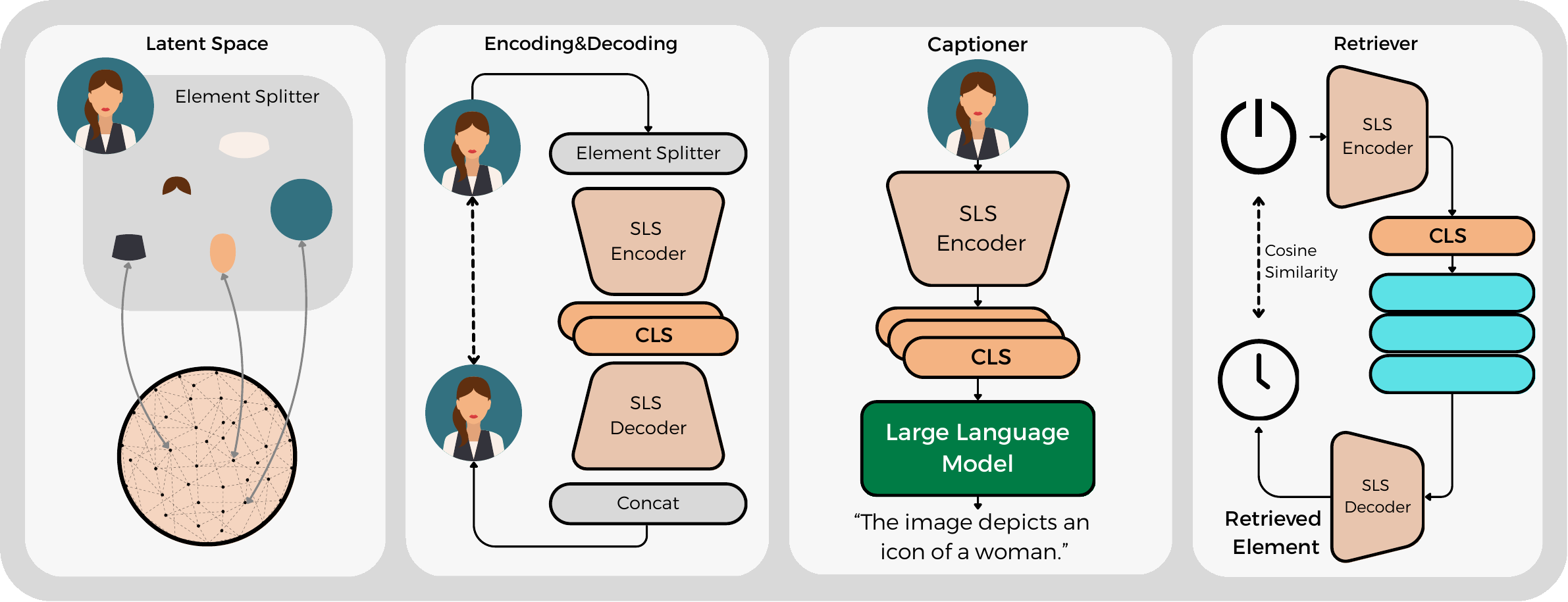}
    \caption{Overview of the proposed approach: from left to right, we show \textit{(i)} \ours path-level latent space, \textit{(ii)} the encoding and decoding framework that learns structured dense representations, and \textit{(iii)} and \textit{(iv)} downstream applications including captioning with a large language model and element-level retrieval via similarity matching.}
    \vspace{-0.8cm}
    \label{fig:teaser}
\end{figure}
We argue that the missing ingredient is a \textit{path-level latent space}: a continuous embedding space where each individual SVG path is mapped to a single dense vector. SVG paths provide a canonical representation for vector graphics, since drawable primitives (\eg, circles, rectangles, ellipses and polygons) can be losslessly converted into path descriptions. As a result, path-level representations form a complete and unified abstraction for SVG content. The analogy to the raster image world is instructive: the introduction of continuous latent spaces for images, most notably through Variational Autoencoders~\cite{kingma2013auto} and their descendants~\cite{rombach2022high, esser2024scaling}, was a watershed moment that unlocked generation, interpolation, retrieval, and downstream conditioning at scale \cite{alves2022variational, xu2023diverse, rafiei2023class, saha2021exploiting}. 
Existing SVG dense encoders~\cite{carlier2020deepsvg, wang2023deepvecfont, polaczek2025neuralsvgimplicitrepresentationtexttovector} are strongly limited in both quality of learned space and sequence representation. Alternatively, traditional raster encoders~\cite{oquab2024dinov, radford2021learning} operate RGB space, discarding the symbolic structure that makes SVGs editable and resolution-independent, and, in the case of raster VAEs, cannot reconstruct the original code from the latent representation, producing features with no path back to valid SVG markup. A path-level latent space should reduce sequence representations from thousands of tokens to a handful of embeddings, preserve both geometric structure and style attributes in a form amenable to vector-space reasoning, and critically decouple the complexity of downstream applications from the length of the underlying SVG markup. Equally important, the latent space must be \emph{invertible}: a dense embedding should be sufficient to faithfully reconstruct the original SVG path, preserving both geometry and style with high fidelity -- a property that any useful vector-domain latent space must provide.

To this end, we introduce \textbf{\ours} (\textbf{S}VG \textbf{L}atent \textbf{S}pace), a Transformer-based autoencoder that learns compact latent representations of individual SVG paths (Figure~\ref{fig:teaser}). A key design choice is the adoption of a data-driven BPE tokenizer trained directly on path corpora, which treats commands, coordinates, and style attributes, such as fill color, stroke width, and opacity -- as a unified token vocabulary. Rather than enforcing rigid command-based boundaries \cite{we2023iconshop, carlier2020deepsvg, wang2023deepvecfont}, BPE discovers sub-word units whose boundaries emerge from corpus statistics, naturally accommodating the heterogeneous mix of geometric and stylistic information that rigid tokenization schemes tend to handle poorly or omit entirely. \ours maps the input sequences to dense vectors via a transformer-based encoder, while a lightweight autoregressive decoder reconstructs the original path from the latent vector with high fidelity. This invertibility is a defining property of \ours: the same dense embedding that enables efficient vector-space reasoning can be decoded back into a valid, style-complete SVG path, something raster encoders such as CLIP~\cite{radford2021learning} or DINOv2~\cite{oquab2024dinov} fundamentally cannot do. \ours operates on single paths rather than full SVG files, reducing average sequence lengths by over two orders of magnitude compared to whole-image LLM-based methods.
Although not explicitly enforced during training, the encoder outputs consistently exhibit a nearly constant $\ell_2$-norm, resulting in latent representations that lie close to a hypersphere of stable radius. We exploit this property by normalizing embeddings at inference time, enabling efficient cosine-similarity search and composition of path embeddings into image-level representations.

We validate \ours on two downstream tasks and analyze the structure of the learned latent space. For SVG image captioning, path embeddings are projected into the token space of pre-trained language models, achieving strong performance across three LLM backbones while reducing training FLOPs by over $150\times$. For path retrieval, \ours enables scalable nearest-neighbor search over hundreds of thousands of paths, outperforming raster-based encoders and prior SVG-specific baselines. Finally, robustness analysis shows that the learned latent space remains stable under both Gaussian noise and angular perturbations.

\noindent The main contributions of this work can be summarized as follows:
\begin{itemize}
    \item We introduce \ours, a scalable and invertible path-level latent space for SVG content that overcomes key limitations of existing approaches, which either lack embedding quality, restrict input flexibility, or omit stylistic attributes.
    \item We show that a data-driven BPE tokenizer applied uniformly to commands, coordinates, and style attributes enables richer and more flexible path representations than rigid command-based alternatives, capturing stylistic variation that prior approaches discard or handle separately.
    \item We demonstrate that \ours learns an invertible latent space that lies on a hypersphere: a single dense vector per path is sufficient to reconstruct the original SVG with high geometric and stylistic fidelity, a property that raster encoders cannot provide.
    \item We show that \ours embeddings generalize across diverse downstream tasks, such as retrieval, captioning, and embedding space analysis, through simple vector-space operations, without task-specific architectural changes.
\end{itemize}
\section{Related Works}
\label{sec:related_works}

\subsection{SVG Representation}
Representing scalable vector graphics remains a central challenge in Computer Vision. Approaches like DeepSVG~\cite{carlier2020deepsvg} introduce hierarchical formulations with fixed per-path control points but omit style attributes. While these methods simplify training and enable structured editing, they sacrifice representational flexibility due to their rigid parameterizations.  
Beyond methods that explicitly model SVGs, a growing body of work demonstrates that neural models trained on tasks involving vector graphics often learn implicit SVG-like representations. For instance, image-to-sequence or sketch generation models, although not constrained to produce valid SVGs, capture underlying geometric and structural patterns that closely resemble vector representations~\cite{ha2017neural}. 

\tit{Image-to-Vector}
Another line of work focuses on recovering vector graphics from raster images. Im2Vec~\cite{reddy2021im2vec} predicts parametric primitives to approximate shapes, while VectorGrimoire~\cite{vectorgrimoire2025cipriano} improves geometric fidelity through richer curve modeling. Other methods, such as DualVector~\cite{liu2023dualvector}, DeepVecFont~\cite{wang2023deepvecfont,lopes2019learned} leverage both image and vector features, focusing specifically on neural representations for font reconstruction and sketch vectorization. These approaches perform well in specialized domains but struggle to generalize to arbitrary SVG content. This also suggests that SVG-like abstractions emerge when models are required to reason about compositional, structured visual content, highlighting the pervasiveness and utility of vector representations in modern deep learning pipelines.

\tit{Large Language Model-based Methods}
Recent text-to-vector models (\eg, IconShop~\cite{we2023iconshop}, vHector~\cite{zini2025vhector}, LLM4SVG~\cite{xing2025empowering}, OmniSVG~\cite{yang2025omnisvg}, SVGen~\cite{wang2025svgen}) and DuetSVG~\cite{zhang2026duetsvg} directly serialize entire SVG images as long token sequences and generate them via large language models. While effective for text-conditioned synthesis, these token-based approaches suffer from significant scalability limitations: even moderately complex graphics require hundreds to thousands of tokens, resulting in sparse, high-dimensional representations that strain context windows and complicate learning.
Rodriguez \etal~\cite{Rodriguez_2025_CVPR} extend this paradigm to image-to-SVG vectorization by treating SVG markup as text sequences, further demonstrating the potential of language models for vector graphics generation. However, the context length problem remains acute -- their approach requires even longer sequences for detailed images, making the representation increasingly sparse and computationally expensive. In contrast, our method addresses this fundamental limitation through compact path-level embeddings that encode geometric and stylistic information in a fixed-size dense representation.

\tit{Hybrid Approaches}
Other works explore intermediate strategies. SVGFusion~\cite{xing2024svgfusion} combines fixed symbolic representations with learnable latent matrices to improve scalability, though it still relies on predefined structural constraints. SuperSVG~\cite{hu2024supersvg} learns SVG representations of superpixel image regions for RGB vectorization, operating at a different granularity than whole-image methods.
Other work, such as NeuralSVG~\cite{polaczek2025neuralsvgimplicitrepresentationtexttovector}, factorizes geometry and color while constraining each path to a fixed number of points.

\subsection{Token-based Encoder-Decoder Architectures}
Sequence-to-sequence architectures have proven effective for learning dense representations across diverse modalities. In machine translation, NLLB~\cite{nllbteam2022languageleftbehindscaling} and SONAR~\cite{Duquenne:2023:sonar} construct unified multilingual embedding spaces that enable cross-lingual transfer through shared semantic representations. Similarly, LCM (Large Concept Models)~\cite{lcm2024} demonstrate that next-embedding prediction can learn structured representations for complex domains. 
Inspired by these successes, our work extends token-based embedding learning to the SVG domain. Unlike prior LLM-based methods that treat entire SVG images as sparse token sequences, we construct a dense path-level embedding space that captures both geometric and stylistic properties in a compact representation, effectively addressing the context length limitations of sequential approaches.
\section{Method}
\label{sec:method}

\subsection{Preliminaries}
We address the problem of learning compact vector representations of SVG paths, which are sequences of discrete drawing commands and associated parameters.
We train a path-specific tokenizer that converts SVG elements into discrete tokens encompassing command types, numeric arguments, and stylistic attributes. To standardize inputs and reduce average sequence length, we adopt the preprocessing procedure introduced in \cite{zini2025vhector}. The tokenizer operates at the individual path level, with compound SVG objects decomposed into independent path sequences.
Unlike prior work~\cite{carlier2020deepsvg,texttovectorwithneuralpathrepresentation}, which relies on specialized geometric encodings, we represent SVG paths as text sequences and train a purely text-based tokenizer that uniformly processes structural and stylistic information. Our tokenizer is based on Byte Pair Encoding (BPE) and is trained on the preprocessed dataset to learn a compact, subword-level vocabulary tailored to this domain. This reduces average sequence length while preserving geometric fidelity. 
Given the resulting token sequence $\mathbf{x} = (x_1, \ldots, x_T)$, our goal is to learn an encoder that maps $\mathbf{x}$ to a dense latent representation, and a decoder that reconstructs the original sequence:
\begin{equation}
    \mathcal{E}(\mathbf{x}) = \mathbf{z}, \quad
    \mathcal{D}(\mathbf{z}) \approx \mathbf{x}, \quad
    \mathbf{x} \in \mathcal{V}^T, \quad
    \mathbf{z} \in \mathbb{R}^d.
\end{equation}
The encoder–decoder architecture follows the standard autoencoder framework for learning compressed representations, enabling downstream applications that benefit from latent vector-space reasoning over structured path data while reducing sparsity.
This approach yields \ours, a SVG autoencoder architecture capable of producing dense representations of complete SVG images and paths while preserving style attributes and managing longer context windows.

 \begin{figure*}[t]
    \centering
    \includegraphics[width=\linewidth]{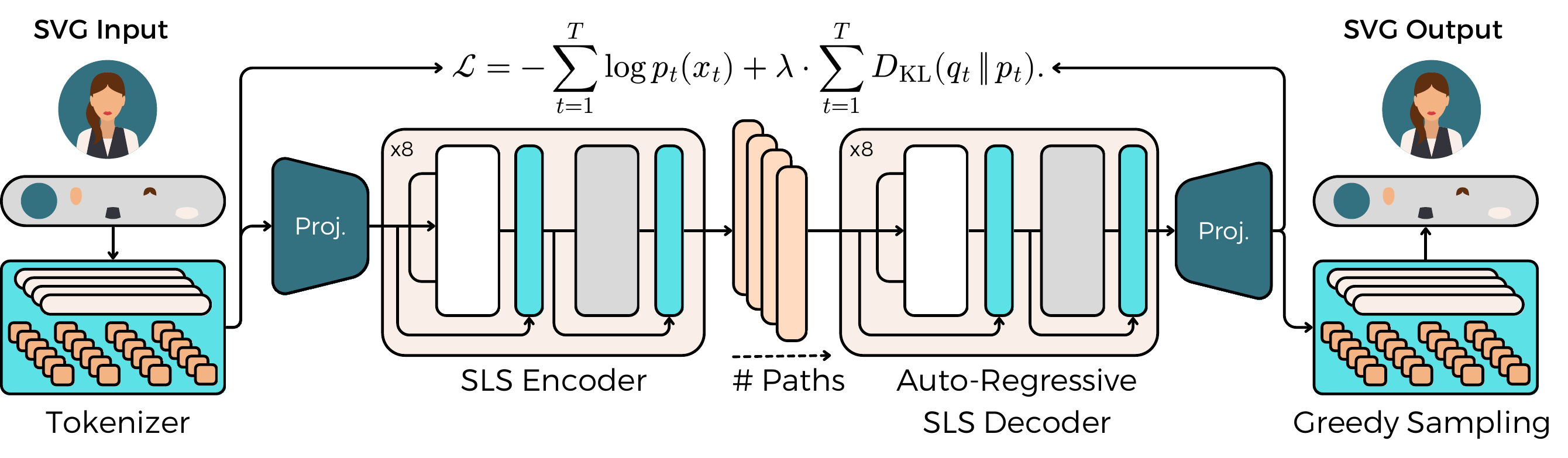}
    \caption{Overview of \ours architecture. SVG paths are encoded into latent representations, with Gaussian noise injection during training for regularization. The autoregressive decoder reconstructs paths through greedy sampling decoding. Embeddings on the unit hypersphere enable efficient downstream applications: similarity-based path retrieval and image captioning via learned projection into language model token space.}
    \vspace{-0.3cm}
    \label{fig:architecture}
\end{figure*}

\subsection{\ours Architecture}
Our proposed approach adopts a Transformer-based encoder-decoder architecture. The overall architecture is illustrated in Figure \ref{fig:architecture}.

\tit{Encoder}
Each SVG path sequence is tokenized and explicitly delimited by special \texttt{<s>} (BOS) and \texttt{</s>} (EOS) tokens, which are part of the vocabulary. Given an input sequence:
\begin{equation}
\mathbf{x} = [\texttt{<s>}, x_1, \ldots, x_T, \texttt{</s>}],
\end{equation}
tokens are mapped to learned embeddings and combined with sinusoidal positional encodings.
The sequence is processed by a stack of Transformer encoder layers.
Instead of prepending a learnable $\text{[CLS]}$ token, we use the final-layer hidden state corresponding to the EOS token as the global sequence representation. Formally, if $\mathbf{H} \in \mathbb{R}^{(T+2)\times d}$ denotes the encoder output, where $T$ is the sequence length and $d$ is the size of the model, the latent embedding is defined as:
\begin{equation}
\mathbf{z} = \mathbf{H}_{\text{EOS}},
\end{equation}
where $\mathbf{H}_{\text{EOS}} \in \mathbb{R}^{d}$ is the hidden state at the EOS position. This representation serves as a dense embedding summarizing the entire input path.

\tit{Decoder}
The decoder reconstructs the original token sequence conditioned on the latent embedding $\mathbf{z} \in \mathbb{R}^{d}$, which matches the decoder model dimension. We inject $\mathbf{z}$ once as a prefix token and apply a causal Transformer over the autoregressive input. A linear language modeling head produces vocabulary logits at each timestep.
Unlike natural language generation, where multiple valid continuations may exist and stochastic sampling strategies can improve diversity~\cite{holtzman2019curious,fan2018hierarchical}, SVG path sequences are deterministic and syntactically strict: each token position admits a single correct value determined by the underlying geometric structure. Therefore, we employ greedy decoding, selecting the token with the highest probability at each step:
\begin{equation}
\hat{x}_t = \argmax_{x \in \mathcal{V}} p(x \mid \mathbf{x}_{<t}, \mathbf{z}),
\end{equation}
where $\mathcal{V}$ denotes the vocabulary. This deterministic strategy ensures syntactically valid path reconstructions and prevents malformed SVG generation.

\tit{Training}
\label{sec:training}
We train the autoencoder end-to-end using token reconstruction loss. To encourage a more robust and well-structured latent representation, we inject Gaussian noise into the latent embedding during training with standard deviation $\sigma = 1.0$, following \cite{carlier2020deepsvg}:
\begin{equation}
\mathbf{z}_{\text{train}} = \mathbf{z} + \epsilon, \quad \epsilon \sim \mathcal{N}(0, \sigma^2 \mathbf{I}).
\end{equation}

Conditioned on the perturbed latent vector, the decoder autoregressively reconstructs the original token sequence. At each timestep, it produces vocabulary logits $\mathbf{y} \in \mathbb{R}^{T \times |\mathcal{V}|}$, which define the predicted token distribution:
\begin{equation}
p_t = \mathrm{softmax}(\mathbf{y}_t).
\end{equation}
The training objective combines the standard cross-entropy loss with a KL divergence term that enforces distributional consistency between the target and predicted token distributions. Specifically, we compute the KL divergence between the target distribution $q_t$ associated with the ground-truth token and the predicted token distribution $p_t$:
\begin{equation}
\mathcal{L}=\underbrace{-\sum_{t=1}^{T} \log p_t(x_t)}_{\mathcal{L}_{\text{CE}}}+\lambda \cdot\underbrace{\sum_{t=1}^{T} D_{\mathrm{KL}}\!\left(q_t \,\|\, p_t\right)}_{\mathcal{L}_{\text{KL}}}.
\end{equation}
Here,  $\lambda$ corresponds to scaling factor of $\mathcal{L}_{\text{KL}}$.
This regularizer stabilizes training by encouraging the predicted distribution to match a sharpened, self-generated target distribution.

\tit{Inference}
At inference time, we observe that the trained encoder produces latent representations with consistent  $\ell_2$ norm with low standard deviation, indicating a well-regularized latent space.
We leverage this property by normalizing the encoder output to unit norm for downstream applications, as
\begin{equation}
\mathbf{z}_{\text{norm}} = \frac{\mathbf{z}}{\|\mathbf{z}\|_2}.
\end{equation}
This normalized representation $\mathbf{z}_{\text{norm}}$ enables efficient vector-space operations (\eg, similarity search, interpolation) in downstream tasks, as all representations lie on the unit hypersphere, and further promotes numerical stability. Importantly, due to the intrinsic constant-norm structure of the learned latent space, this normalization does not discard meaningful magnitude information, thereby avoiding information loss. When reconstruction is required, we rescale by the empirical mean before feeding to the decoder:
\begin{equation}
\mathbf{z}_{\text{decoder}} = \mu_{\|\mathbf{z}\|} \cdot \mathbf{z}_{\text{norm}},
\end{equation}
where  $\mu_{\|\mathbf{z}\|}$ indicates the mean $\ell_2$ norm characteristic of the model.
This decoupling between the normalized encoder output and the rescaled decoder input provides flexibility: downstream applications can reason with unit-norm vectors while the decoder receives representations in the expected magnitude range. 

\subsection{Downstream Applications}
The normalized latent representations $\mathbf{z}_{\text{norm}} \in \mathbb{R}^d$ produced by our encoder enable various downstream applications that benefit from compact, semantically meaningful vector representations of SVG paths and images.

\tit{Path Retrieval}
Given a query path encoded as $\mathbf{z}_q$, we retrieve the most similar paths from a database by computing cosine similarity in the normalized embedding space:
\begin{equation}
\text{sim}(\mathbf{z}_q, \mathbf{z}_i) = \frac{\mathbf{z}_q^\top \mathbf{z}_i}{\|\mathbf{z}_q\| \|\mathbf{z}_i\|}.
\end{equation}
Since all embeddings lie on the unit hypersphere, cosine similarity reduces to the dot product enabling efficient retrieval with favoring nearest neighbor methods. This capability is particularly valuable for codebook-based applications and content-based search in large SVG databases.

\tit{SVG Image Captioning}
We extend our approach to generate natural language descriptions of complete SVG images. Given an SVG image composed of multiple paths, we encode each path independently to obtain a sequence of path embeddings $\{\mathbf{z}_1, \ldots, \mathbf{z}_N\}$. These embeddings are projected into the token embedding space of a pre-trained language model via a learned linear transformation:
\begin{equation}
\mathbf{e}_i = \mathbf{W}_{\text{proj}} \mathbf{z}_i ,
\end{equation}
where $\mathbf{W}_{\text{proj}} \in \mathbb{R}^{d_{\text{LLM}} \times d}$  is a  learned matrix of parameters. The projected embeddings are prepended to the language model's input as prefix tokens, conditioning the autoregressive generation on the visual content:
\begin{equation}
p(\text{caption} \mid \text{SVG}) = \prod_{t=1}^{T_{\text{cap}}} p(w_t \mid w_{<t}, \mathbf{e}_1, \ldots, \mathbf{e}_N),
\end{equation}
where $w_t$ denotes the $t$-th word in the caption. This formulation treats SVG path embeddings as visual tokens, enabling the language model to ground generation in the geometric and stylistic properties encoded by \ours.
\section{Experiments}
\label{sec:experiments}

\subsection{Experimental setup}

\tit{Dataset}
\label{sec:dataset}
To train and evaluate our model, we utilize a combination of publicly available SVG datasets, namely StarVector~\cite{Rodriguez_2025_CVPR}, HeisenVec~\cite{zini2025vhector}, ColorSVG~\cite{chen2024svgbuilder}, and SVGX-Core~\cite{xing2025empowering}. In addition to SVG images, all datasets include captions automatically generated by a multimodal large language model (MLLM).
Since our approach focuses on SVG path representations, we standardize and preprocess all data following the filtering procedure proposed in~\cite{zini2025vhector}. Specifically, we filter out paths exceeding 1024 tokens to ensure computational efficiency and maintain consistency across the training set, while leaving sufficient headroom for future context length extensions.
This preprocessing pipeline yields a large-scale curated dataset comprising approximately 1.5M images for training, 31k for validation, and 16k for testing. At the path level, the dataset contains 25M training paths, 500k validation paths, and 250k test paths, providing sufficient data for robust SVG path encoding and generation tasks.

\tit{Training Details}
We train \ours from scratch for 480k steps using a batch size of 192 and a learning rate of $1\times10^{-4}$ with 6k warmup steps. The encoder and decoder share the same architecture with a hidden size of 1024, 8 attention heads, and an MLP ratio of 2, the overall number of parameters for both encoder and decoder is 135M. Training is performed on 4 NVIDIA A100 GPUs.

\subsection{Reconstruction task}
To validate the quality of \ours, we measured reconstruction performance at two levels: (i) \textit{image level}, comparing original and reconstructed images using classical Computer Vision metrics (MSE-similarity, SSIM~\cite{1284395}, LPIPS~\cite{zhang2018unreasonable}, DINOv2-Similarity~\cite{oquab2024dinov}), and (ii) \textit{path level}, analyzing token distribution with BLEU~\cite{papineni2002bleu} and METEOR~\cite{banerjee2005meteor}, and command coherency by computing the mean intersection over union (mIoU) of reconstructed paths over the original, without accounting for stylistic differences.

Table \ref{tab:encoder_comparison_compact} shows \ours performance against that of DeepSVG \cite{carlier2020deepsvg} in terms of reconstruction capabilities. As DeepSVG is trained to embed entire SVG images rather than single paths, we compare against it only at the image level, evaluating its capabilities both in a zero-shot setting and after retraining it on our dataset. As can be seen, \ours outperforms DeepSVG by a significant margin on image-level reconstruction, even when retraining it, while respecting their limitation on maximum number of commands per path, which is one the its main limitations.

\begin{figure}[t]
    \centering
    \includegraphics[width=0.97\linewidth]{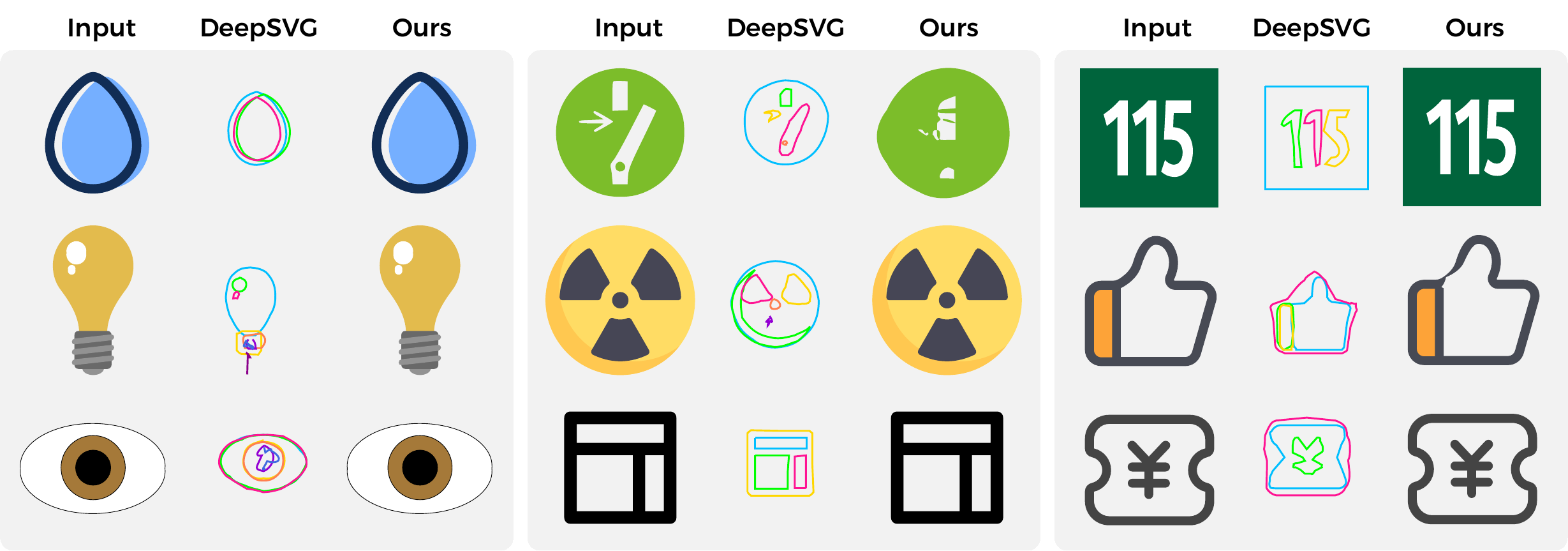}
    \caption{Qualitative results on reconstruction of input image.}
    \vspace{-.3cm}
    \label{fig:qualitatives_reconstruction}
\vspace{-.3cm}
\end{figure}

\noindent Further, given that \ours produces path embeddings with constant norm, we also investigate whether magnitude information can instead be encoded during training through alternative mechanisms. We experimented with three techniques: (i) $\ell_2$ regularization to minimize the CLS norm, (ii) multiplying the CLS token by a learnable logit scale to encode magnitude information in the scale factor rather than the embedding norm, and (iii) combining both approaches. Table \ref{tab:encoder_comparison_compact} shows that these techniques do not learn better representations compared to our final approach, confirming that constant $\ell_2$ norm peculiarity emerges spontaneously.
Figure \ref{fig:qualitatives_reconstruction} provides qualitative evidence that \ours consistently outperforms DeepSVG in reconstruction quality. This improvement is attributed to our model end-to-end design, which jointly processes style attributes and command/argument tokens at the input level, resulting in a latent representation that captures both geometric and visual properties.

\begin{table*}[t]
\centering
\caption{Reconstruction performance at image and path levels.}
\vspace{-.3cm}
\label{tab:encoder_comparison_compact}
\footnotesize
\setlength{\tabcolsep}{0.7em}
\resizebox{.97\textwidth}{!}{
\begin{tabular}{lccccccc}
\toprule
\multirow{2}{*}{\textbf{Method}} 
& \multicolumn{4}{c}{\textbf{Image}} 
& \multicolumn{3}{c}{\textbf{Path}} \\
\cmidrule(lr){2-5} \cmidrule(lr){6-8}
& MSE$\uparrow$ 
& SSIM$\uparrow$ 
& LPIPS$\downarrow$ 
& DINO$\uparrow$ 
& mIoU$\uparrow$ 
& BLEU$_5\uparrow$ 
& MET$\uparrow$ \\
\midrule

\rowcolor{gray!10}\textit{Baseline} &&&&&&&\\
DeepSVG~\cite{carlier2020deepsvg}
& 76.03 & 67.85 & 53.99 & 49.86 & -- & -- & -- \\
\quad retrained
& 76.05 & 67.85 & 53.31 & 57.95 & -- & -- & -- \\

\midrule
\rowcolor{gray!10}\textit{Ours} &&&&&&&\\

\quad w/ $\ell_2$
& 86.51 & 81.05 & 28.63
& 72.41 & 66.79
& 62.05 & 90.02 \\

\quad w/ logit
& 71.98 & 58.63 & 62.38
& 36.83 & 5.63
& \textbf{98.91} & 77.45 \\

\quad w/ logit+$\ell_2$
& 71.72 & 59.65 & 63.68
& 36.98 & 5.45
& 97.06 & 71.73 \\

\midrule
\rowcolor{OurColor}\textbf{\ours (BPE)}
& \textbf{90.83} & \textbf{87.26} & \textbf{18.79}
& \textbf{80.85} & \textbf{78.92}
& 96.21 & \textbf{97.07} \\

\bottomrule
\end{tabular}
}
\vspace{-.3cm}
\end{table*}

\tit{Ablation studies} Table \ref{tab:ablation_studies} further analyzes architectural and training choices, including projection dimensionality, pooling strategy (ViT-like $\text{[CLS]}$ token vs. sequence modeling using the EOS representation), KL regularization, alternative tokenization schemes, and a block-wise image decomposition strategy. Results confirm that data-driven BPE tokenization and EOS-based pooling yield the best trade-off across reconstruction metrics, while command-based tokenizers significantly degrade performance. To assess whether fixed-length chunking could approximate path-level modeling, we experimented with multiple block sizes, consistently observing inferior performance compared to semantically coherent path decomposition. We find out that complete paths preserves both geometric structure and stylistic consistency better than fixed block chunking.

\subsection{SVG Image Captioning}
\tit{Experimental Setup}
We then move to evaluating the quality of \ours embeddings through SVG-to-text generation. Following standard vision-language practices, we train lightweight projection layers that map \ours path embeddings into the input space of pre-trained language models. We experiment with three model scales representing different efficiency-capability trade-offs: Qwen3 0.6B~\cite{yang2025qwen3technicalreport}, Llama 3.2 1B~\cite{grattafiori2024llama3herdmodels}, and Gemma 2 2B~\cite{team2024gemma}. 

The projection layer and language model are jointly trained end-to-end until convergence, allowing both components to adapt to the visual-language alignment task. All captioning models use a learning rate of $1\times10^{-4}$ and batch size of 384. Additional training details are provided in the supplementary materials.

\begin{table*}[t]
\centering
\caption{Architectural and training ablation studies. We evaluate the effect of \textit{(i)} embedding projection size, \textit{(ii)} training loss and pooling strategy (CLS vs. EOS, with/without KL), \textit{(iii)} command-based tokenization approaches, and \textit{(iv)} a block-wise image decomposition strategy instead of path-level modeling.}
\vspace{-.3cm}
\label{tab:ablation_studies}
\footnotesize
\setlength{\tabcolsep}{.6em}
\resizebox{.97\textwidth}{!}{
\begin{tabular}{lccccccc}
\toprule
\multirow{2}{*}{\textbf{Method}} 
& \multicolumn{4}{c}{\textbf{Image}} 
& \multicolumn{3}{c}{\textbf{Path}} \\
\cmidrule(lr){2-5} \cmidrule(lr){6-8}
& MSE$\uparrow$ 
& SSIM$\uparrow$ 
& LPIPS$\downarrow$ 
& DINO$\uparrow$ 
& mIoU$\uparrow$ 
& BLEU$_5\uparrow$ 
& MET.$\uparrow$ \\
\midrule

\rowcolor{gray!10}\textit{Projection size} &&&&&&&\\
 512   & 79.37 & 71.44 & 51.47 & 47.99 & 26.16 & 46.95 & 65.82 \\
 768   & 79.69 & 72.03 & 46.76 & 52.48 & 43.91 & 66.61 & 78.52 \\

\midrule
\rowcolor{gray!10}\textit{Pooling and training} &&&&&&&\\
LAST & 90.77 & 87.13 & 19.27 & \textbf{81.08} & \textbf{79.17} & 96.17 & 97.04 \\
CLS & 90.16 & 86.26 & 20.02 & 79.72 & 78.87 & 96.03 & 96.93 \\
$\triangleright$ w/ KL Loss & 90.17 & 86.34 & 19.87 & 79.97 & 79.05 & 96.13 & 96.97 \\

\midrule
\rowcolor{gray!10}\textit{Tokenizers} &&&&&&&\\
 DeepSVG-like~\cite{carlier2020deepsvg}  & 75.60 & 67.08 & 47.73 & 53.71 & 40.58 & 5.33 & 23.35 \\
Iconshop-like~\cite{we2023iconshop}  & 79.26 & 67.61 & 46.01 & 58.34 & 21.86 & 1.68 & 16.57 \\

\midrule
\rowcolor{gray!10}\textit{Block size} &&&&&&&\\
 512 tokens & 88.77 & 83.73 & 23.99 & 76.22 & 64.31 & 63.63 & 74.51 \\
 1024 tokens & 89.21 & 84.13 & 24.06 & 76.84 & 61.33 & 61.46 & 70.10 \\

\midrule
\rowcolor{OurColor}\textbf{\ours (Ours)}
& \textbf{90.83} & \textbf{87.26} & \textbf{18.79}
& 80.85 & 78.92
& \textbf{96.21} & \textbf{97.07} \\
\bottomrule
\end{tabular}
}
\vspace{-.6cm}
\end{table*}

\tit{Evaluation Metrics}
We evaluate caption quality using complementary metrics. ROUGE~\cite{lin2004automatic} emphasizes recall for longer texts, while BLEU~\cite{papineni2002bleu} measures n-gram overlap with reference captions. METEOR~\cite{banerjee2005meteor} accounts for synonyms and paraphrasing. CLIP-Score~\cite{hessel2021clipscore} quantifies the semantic alignment between the generated caption and the rasterized SVG image using CLIP embeddings.

\tit{Results}
To evaluate the semantic quality of our embedding space, we trained a set of LLMs to generate captions from SVG embeddings. Table \ref{tab:model_comparison} shows different encoders across three LLM backbones, and compares \ours to DeepSVG~\cite{carlier2020deepsvg}, the plain XML code, and CLIP~\cite{radford2021learning}. Here, CLIP~\cite{radford2021learning} serves as an upper-bound reference: while it produces semantically rich embeddings suitable for captioning, it cannot decode back to SVG, operating in a fundamentally different domain.

\begin{figure}[tb]
    \centering
    \begin{minipage}[h]{0.49\linewidth}
        \centering
        \includegraphics[width=\linewidth]{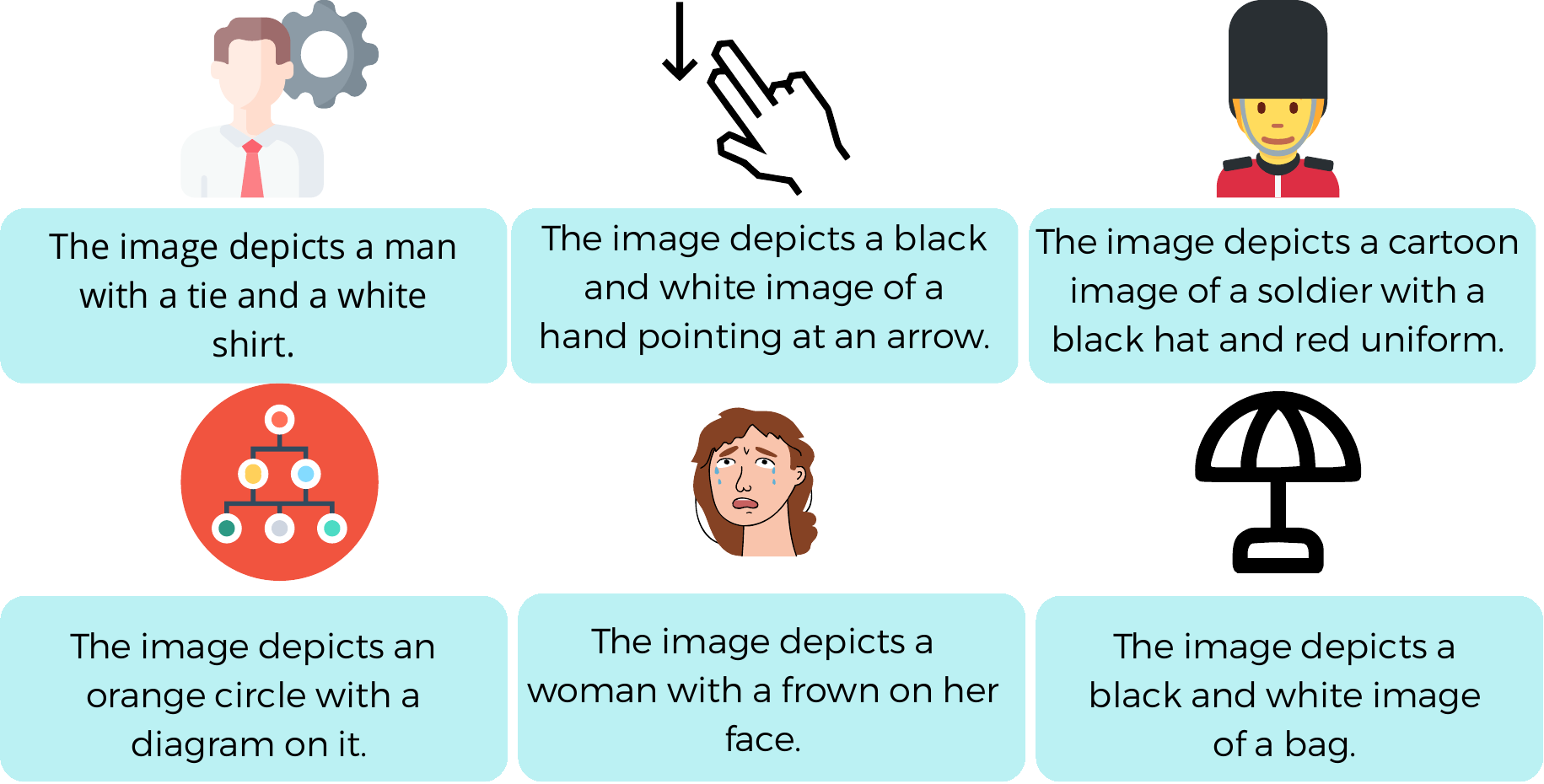}
        \vspace{-.6cm}
        \caption{Qualitative captioning samples of Llama 3.2 1B~\cite{grattafiori2024llama3herdmodels} as Large Language Model using \ours as SVG image encoder.}
        \label{fig:qualitatives_captioning}
    \end{minipage}
    \hfill
    \begin{minipage}[h]{0.49\linewidth}
    \centering
        \includegraphics[width=\linewidth]{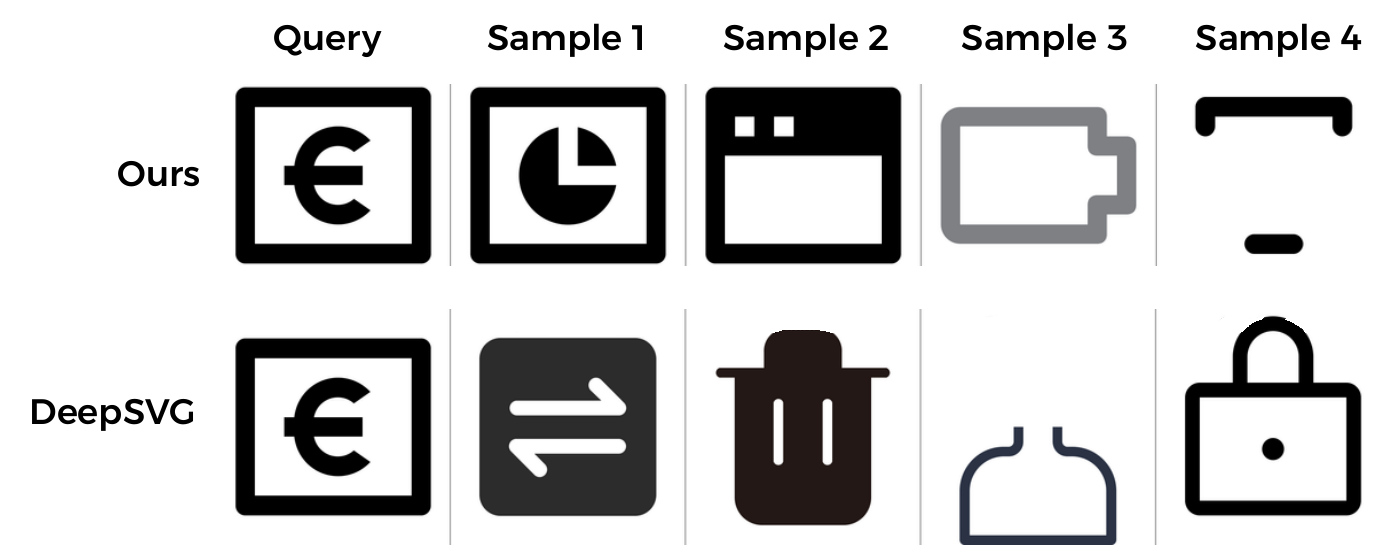}
        \caption{Qualitative path retrieval results. Query paths (left) and top-ranked retrievals ordered by cosine similarity for \ours (Ours) and DeepSVG~\cite{carlier2020deepsvg}.}
        \label{fig:qualitatives_retrieval}
    \end{minipage}
    \vspace{-0.3cm}
\end{figure}

As can be seen, encoding SVG paths directly as XML text tokens performs poorly despite preserving complete information, as the extreme sequence lengths required make learning prohibitive. \ours approach achieves significantly better performance (+4.2 CLIP-Score, +29.01 $\text{BLEU}_{5}$) using dense embeddings, validating the effectiveness of our learned compression for downstream tasks. Further, \ours consistently outperforms DeepSVG across all metrics and LLMs, with substantial average gains in CLIP-Score (+2.82), $\text{BLEU}_{5}$ (+19.19), and METEOR (+24.63), demonstrating superior semantic richness while maintaining invertibility. Qualitative results for SVG captioning are presented in Figure \ref{fig:qualitatives_captioning}, where \ours serves as the visual encoder paired with Llama 3.2 1B as the language model, demonstrating strong caption quality.

\begin{table*}[t]
\caption{SVG captioning performance with different encoders and language models. \ours surpasses invertible baselines (DeepSVG, XML) across all evaluation metrics, demonstrating superior semantic quality of learned embeddings. CLIP (gray) represents a non-invertible reference.}
\vspace{-.2cm}
\centering
\setlength{\tabcolsep}{0.6em}
\resizebox{.97\textwidth}{!}{
\begin{tabular}{lclccccc}
\toprule
\textbf{Encoder} & \textbf{SVG} & \multirow{1}{*}{\textbf{Backbone}} & \textbf{T2I} $\uparrow$ &  \multirow{1}{*}{\textbf{$\text{BLEU}_{5}$}$\uparrow$} & \multirow{1}{*}{\textbf{METEOR}$\uparrow$}  & \multirow{1}{*}{\textbf{$\text{ROUGE}_{1}$}$\uparrow$}  & \multirow{1}{*}{\textbf{$\text{ROUGE}_{2}$}$\uparrow$}   \\

\midrule

\rowcolor{gray!10} &  & Qwen3 0.6B~\cite{yang2025qwen3technicalreport} & 29.48 &  49.44 & 69.12 & 75.23 & 63.48\\
\rowcolor{gray!10}CLIP~\cite{radford2021learning}& \xmark &Gemma2 2B~\cite{team2024gemma} & 29.03 &  24.78 & 53.72 & 55.70& 41.40\\
\rowcolor{gray!10} &  & Llama 3.2 1B~\cite{grattafiori2024llama3herdmodels} & 29.57 &  54.67& 71.87 & 77.39& 67.18 \\

\cmidrule{1-8}
\cmidrule{1-8}

& & Qwen3 0.6B~\cite{yang2025qwen3technicalreport} & 21.95 &  10.38 & 32.14 & 38.84 & 23.49\\
XML & \cmark& Gemma2 2B~\cite{team2024gemma} & 22.73 &  12.62 & 34.48 & 44.31& 26.67\\
& & Llama 3.2 1B~\cite{grattafiori2024llama3herdmodels} & 23.99 &  12.31 & 32.45 & 43.25 & 28.13\\

\midrule

& &Qwen3 0.6B~\cite{yang2025qwen3technicalreport} & 25.42 &  27.30 & 40.55 & 48.96 & 37.02\\
DeepSVG~\cite{carlier2020deepsvg}& \cmark& Gemma2 2B~\cite{team2024gemma} & 25.39 & 18.71 & 37.07 & 44.95& 27.49\\
& & Llama 3.2 1B~\cite{grattafiori2024llama3herdmodels} & 23.25 &  18.77 & 35.82 & 44.77& 27.22\\
\cmidrule{1-8}

& &Qwen3 0.6B~\cite{yang2025qwen3technicalreport} & \textbf{27.69} & 45.60 & 64.95 & 70.43 & 59.06 \\
\textbf{\ours (Ours)} & \cmark & Gemma2 2B~\cite{team2024gemma} & 27.15 &  29.86 & 56.57 & 59.18 & 45.47 \\

& & Llama 3.2 1B~\cite{grattafiori2024llama3herdmodels} & \textbf{27.69} &  \textbf{46.89} & \textbf{65.80} & \textbf{71.29} & \textbf{60.04} \\

\bottomrule
\end{tabular}
}
\label{tab:model_comparison}
\vspace{-.2cm}
\end{table*}

\tit{Computational Efficiency}
Table~\ref{tab:comp_spe} reports the computational cost of SVG captioning with and without the \ours encoder. Replacing raw XML token sequences with our dense path embeddings reduces the average context length from 2432 to just 15.56 tokens, yielding a $156\times$ speedup in context processing and a $167\times$ reduction in training TFLOPs. This gain stems directly from our intrinsic learned compression: rather than feeding thousands of XML tokens per path to the language model, a single compact embedding captures the full geometric and stylistic content of each path. Crucially, this efficiency does not come at the cost of quality, \ours outperforms both DeepSVG and the raw XML baseline across all captioning metrics. While DeepSVG similarly avoids rasterization and could in principle offer comparable efficiency benefits, it lacks the stylistic and semantic richness encoded by \ours, which translates into consistently lower caption quality, showing its embedding limitations. These results highlight that compact path-level embeddings not only improve scalability, but also fundamentally decouple downstream computational cost from the length and complexity of the underlying SVG markup.

\begin{table}[t]
\centering
\caption{Computational analysis of SVG image captioning with and without \ours. Our compact path-level representations reduce the average context length by $156\times$ and the training cost by over $167\times$ TFLOPs across three LLM backbones.}
\vspace{-.2cm}
\setlength{\tabcolsep}{1em}
\resizebox{0.99\linewidth}{!}{
\begin{tabular}{llccccc}
\toprule
 &  & \smallhead{Llama3 1B} & \smallhead{Qwen3 0.6B} & \smallhead{Gemma2 2B} & \multirow{1}{*}{\smallhead{Avg.}} & \multirow{1}{*}{\smallhead{Speedup}} \\

\midrule
\textbf{Context} & w/o \ours      & 1636.55  & 2803.26  & 2856.27  & 2432.02  & \textbf{\multirow{2}{*}{156.29}} \\
\textbf{length}  & w/ \ours       & 15.56 & 15.56 & 15.56 & 15.56 \\
\midrule 
\midrule
\textbf{Training} & w/o \ours   & 9.82  & 10.09  & 34.27  & 18.06  & \textbf{\multirow{2}{*}{167.22}} \\
\textbf{TFLOPs} & w/ \ours   & 0.09 & 0.05 & 0.18 & 0.108 &  \\
\bottomrule
\end{tabular}
}
\vspace{-.2cm}
\label{tab:comp_spe}
\end{table}

\begin{table}[t]
\caption{Path retrieval performance on the SVG test set. \ours outperforms SVG-native encoders by retrieving  more faithful SVG paths.}
\vspace{-.2cm}
\centering
\small
\resizebox{.99\columnwidth}{!}{
\setlength{\tabcolsep}{1.2em}
\begin{tabular}{lccccc}
\toprule
\textbf{Model} & \smallhead{Raster} & \smallhead{MSE-Sim $\uparrow$} & \smallhead{LAB Dist. $\downarrow$} & \smallhead{$\text{BLEU}_{5}\uparrow$} & \smallhead{METEOR $\uparrow$} \\
\midrule
CLIP-B~\cite{radford2021learning}& \cmark & 98.67 & \underline{11.66} & 10.58 & 34.12 \\
DINOv2-B~\cite{oquab2024dinov} & \cmark & \underline{98.86} & 14.68 & 10.77 & 34.55 \\
\midrule
\midrule
DeepSVG~\cite{carlier2020deepsvg}& \xmark & 81.45 & 32.06 & 12.89 & 43.27 \\
\rowcolor{OurColor}\textbf{\ours (Ours)} & \xmark & \textbf{89.95} & \textbf{25.27} & \underline{\textbf{20.91}} & \underline{\textbf{47.78}} \\
\bottomrule
\end{tabular}
\label{tab:retrieval_metrics}
}
\vspace{-.3cm}
\end{table}
\subsection{Path Retrieval}
\tit{Experimental Setup}
We evaluate the path retrieval performance of different encoders on a large-scale benchmark. Specifically, with respect to the constraint posed by DeepSVG, we randomly choose 2.5k query paths and 266k document paths from our validation dataset. Each path was then independently embedded using its respective encoder, and all embeddings were $\ell_2$-normalized before computing pairwise similarities. We then calculated the cosine similarity matrix between query and database embeddings.
For each query, we select the top-1 retrieved path (i.e., the database path with highest cosine similarity). Since DeepSVG and \ours can decode embeddings back to SVG, while raster-based encoders such as DINOv2~\cite{oquab2024dinov} and CLIP~\cite{radford2021learning} cannot, we recover the corresponding original SVG paths using the retrieved embedding indices for fair comparison.
Retrieval quality was assessed through both visual and syntactic metrics. Visual similarity was measured using MSE-Sim and distance in CIE-LAB space, which captures differences in the color attributes. Syntactic fidelity was evaluated using $\text{BLEU}_{5}$ and METEOR, quantifying token-level pairs overlap.

\tit{Results}
Table \ref{tab:retrieval_metrics} reports path retrieval performance compared to DeepSVG~\cite{carlier2020deepsvg}, CLIP~\cite{lopes2019learned}, and DINOv2~\cite{oquab2024dinov}. Among SVG-based methods, \ours achieves the best overall performance, significantly outperforming DeepSVG in both visual and syntactic metrics. In particular, \ours attains higher MSE-Sim, lower LAB distance, and improved $BLEU_5$ and $METEOR$, indicating better visual consistency and stronger structural fidelity. While raster-based encoders (CLIP, DINOv2) achieve higher visual similarity due to image-level supervision, they are less sensitive to vector structure. In contrast, SVG-based models better capture geometric and syntactic variations. As illustrated in Figure~\ref{fig:qualitatives_retrieval}, \ours consistently retrieves more structurally similar paths, establishing it as the most effective encoder for vector path retrieval.

 \begin{figure}[t]
     \centering
     \includegraphics[width=1\linewidth]{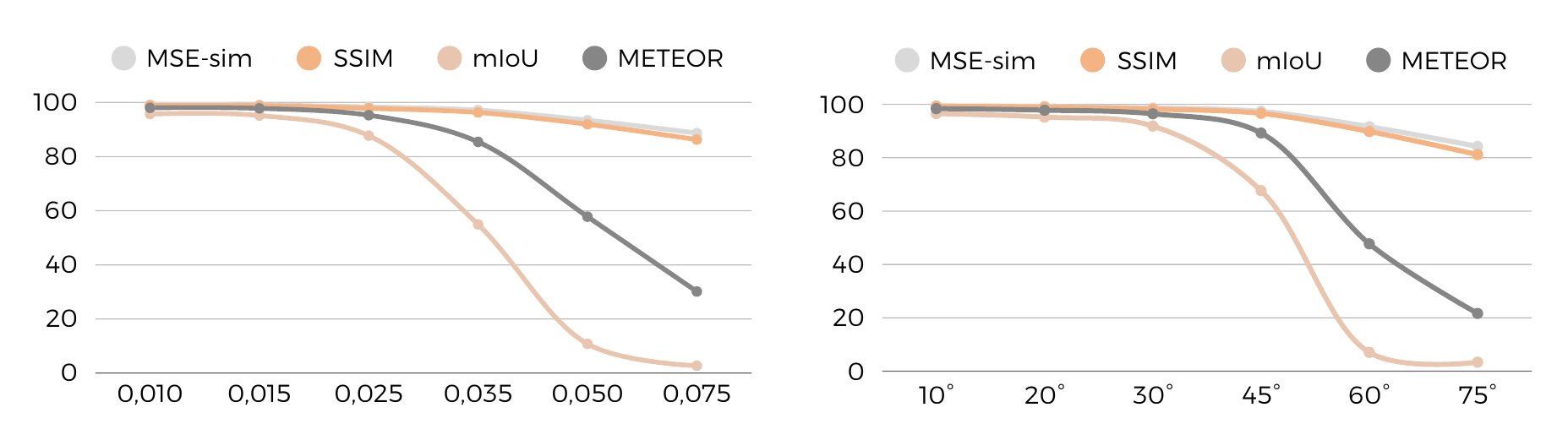}
     \vspace{-.4cm}
     \caption{
 Reconstruction performance of \ours under varying noise levels. Additive Gaussian noise (left) and rotational perturbations applied to input vectors (right). }
     \label{fig:gaussian}
     \vspace{-.3cm}
 \end{figure}

\subsection{Space Robustness}

To assess the stability of the learned embedding space, we systematically evaluate the effect of controlled perturbations applied directly to the latent representations. Two complementary regimes were considered: Gaussian noise and angular perturbations. Gaussian additive noise was introduced as
\begin{equation}
\tilde{\mathbf{z}} = \mathbf{z} + \epsilon, \quad \epsilon \sim \mathcal{N}(0, \sigma^2 \mathbf{I}),
\end{equation}
with $\sigma$ controlling the noise magnitude.  
In the angular regime, we perturbed the embeddings along the tangent space of the unit hypersphere, rotating them by increasing angular offsets $\theta$, as follows:
\begin{equation}
\begin{gathered}
\tilde{\mathbf{z}} = \|\mathbf{z}\| \bigl[ 
\cos(\Delta \theta)\,\hat{\mathbf{z}} 
+ \sin(\Delta \theta)\,\hat{\mathbf{y}} 
\bigr], \\
\text{where }  
\hat{\mathbf{z}} = \frac{\mathbf{z}}{\|\mathbf{z}\|}
\quad 
\hat{\mathbf{y}} = \frac{\mathbf{r} - (\mathbf{r}^\top \hat{\mathbf{z}})\hat{\mathbf{z}}}
{\|\mathbf{r} - (\mathbf{r}^\top \hat{\mathbf{z}})\hat{\mathbf{z}}\|},
\quad
\mathbf{r} \sim \mathcal{N}(0, I_d);
\end{gathered}
\end{equation}
this simulates directional displacement within the latent manifold while preserving embedding norm.
For each perturbed embedding $\tilde{\mathbf{z}}$, we decoded the corresponding SVG path with \ours decoder and compared it to the original using both visual and structural metrics. This setup provides a fine-grained view of how local perturbations in latent space affect geometric and syntactic consistency in the reconstructed paths.
Results are visually reported in Figure~\ref{fig:gaussian}, where we notice that the space learned with \ours is robust to rotations up to 30 degrees, and up to $\sigma = 0.02$ when applying Gaussian noise on $\ell_2$-norm embeddings. This further attests the robustness of the embedding space.

\subsection{Generalization Beyond Path-Level Reconstruction}

We investigate whether the learned latent space generalizes across datasets and naturally extends to image-level representations. To assess cross-dataset generalization, we evaluate \ours on the UniSVG dataset~\cite{li2025unisvg}, which is not included in the training mixture. 
As reported in Table~\ref{tab:ood}, despite the domain shift, \ours exhibits
\begin{wraptable}{r}{0.50\columnwidth}
\vspace{-0.3cm}
\centering

\caption{\ours performance in in-domain (ID) and out-of-domain (OOD) settings.}
\vspace{-0cm}
\resizebox{\linewidth}{!}{
\setlength{\tabcolsep}{0.3em}
\begin{tabular}{lcccc}
\toprule
\textbf{Model} &
\smallhead{MSE $\uparrow$} &
\smallhead{DINO $\uparrow$} &
\smallhead{LPIPS $\downarrow$} &
\smallhead{SSIM $\uparrow$} \\
\midrule
ID     & 90.83 & 80.85 & 18.79 & 87.26\\
OOD & 86.67 & 76.12 & 30.18 & 78.44\\
\bottomrule
\label{tab:ood}
\end{tabular}
}

\vspace{0cm}

\caption{Multiple path representation.}
\vspace{-0cm}
\resizebox{\linewidth}{!}{
\setlength{\tabcolsep}{.8em}
\begin{tabular}{lccc}
\toprule
\textbf{Model} &
\smallhead{Acc. $\uparrow$} &
\smallhead{Prec. $\uparrow$} &
\smallhead{Rec. $\uparrow$}\\
\midrule
DeepSVG & 15.36 & 17.89 & 15.02\\
SLS & \textbf{29.98} & \textbf{30.93} & \textbf{29.62}\\
\bottomrule
\label{tab:multiple-path}
\end{tabular}
}

\vspace{-0.5cm}

\end{wraptable}
only a moderate degradation with respect to the in-domain evaluation, while maintaining high perceptual and structural reconstruction quality. 
These results indicate that the learned latent representations capture general properties of SVG paths rather than overfitting to the training distribution.
Furthermore, although \ours is trained exclusively on individual path reconstruction, complete SVG images can be represented by aggregating the embeddings of their constituent paths. To evaluate these image-level representations, we train a lightweight Transformer classifier on the 20 most frequent classes of the ColorSVG~\cite{chen2024svgbuilder} dataset, restricting each image to at most eight paths. As shown in Table~\ref{tab:multiple-path}, \ours nearly doubles the classification performance of DeepSVG across accuracy, precision, and recall, demonstrating that the learned path embeddings effectively compose into expressive image-level representations that generalize well to downstream image-level tasks.
\section{Conclusion}
\label{sec:conclusion}

We presented \ours, a Transformer-based autoencoder that learns compact, invertible latent representations of individual SVG paths. By treating path commands, coordinates, and style attributes as a unified token vocabulary through data-driven BPE tokenization, \ours produces dense embeddings that capture both geometric and stylistic information in a fixed-size vector — reducing sequence lengths by orders of magnitude compared to full-SVG token-based approaches. The learned space exhibits robustness to latent perturbations and generalizes to diverse downstream tasks, such as retrieval and captioning, through simple vector-space operations. By enabling compact, semantically structured representations of individual paths, we believe \ours lays the groundwork for future research in vector-native modeling, bridging symbolic vector representations with modern representation learning and large-scale foundation models.

\section*{Acknowledgements}
This work has been conducted under a research grant co-funded by Doxee S.p.A. and supported by the EU Horizon project “ELLIOT - European Large Open Multi-Modal Foundation Models For Robust Generalization On Arbitrary Data Streams” (No. 101214398) and by the EU Horizon projects “ELIAS - European Lighthouse of AI for Sustainability” (No. 101120237). We further acknowledge the CINECA award, under the ISCRA initiative, for the availability of high-performance computing resources.

\bibliographystyle{splncs04}
\bibliography{main}

\newpage

\clearpage


\section*{Supplementary material}

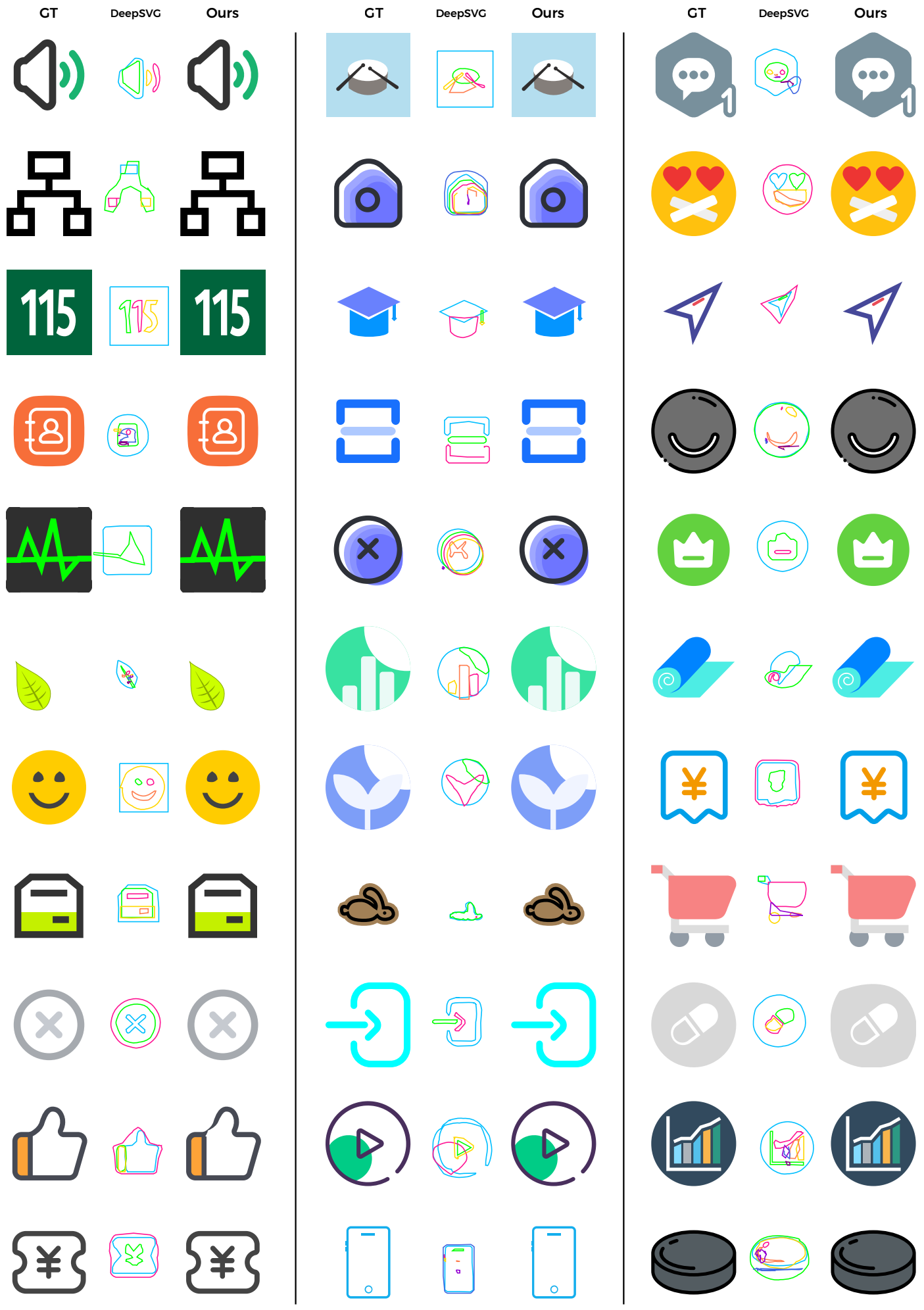


**Fig. 1:** Qualitative results on image reconstruction. Each separated column shows three samples from left to right: ground truth (GT) SVG images, DeepSVG reconstructions, and SLS (Ours) reconstructions.

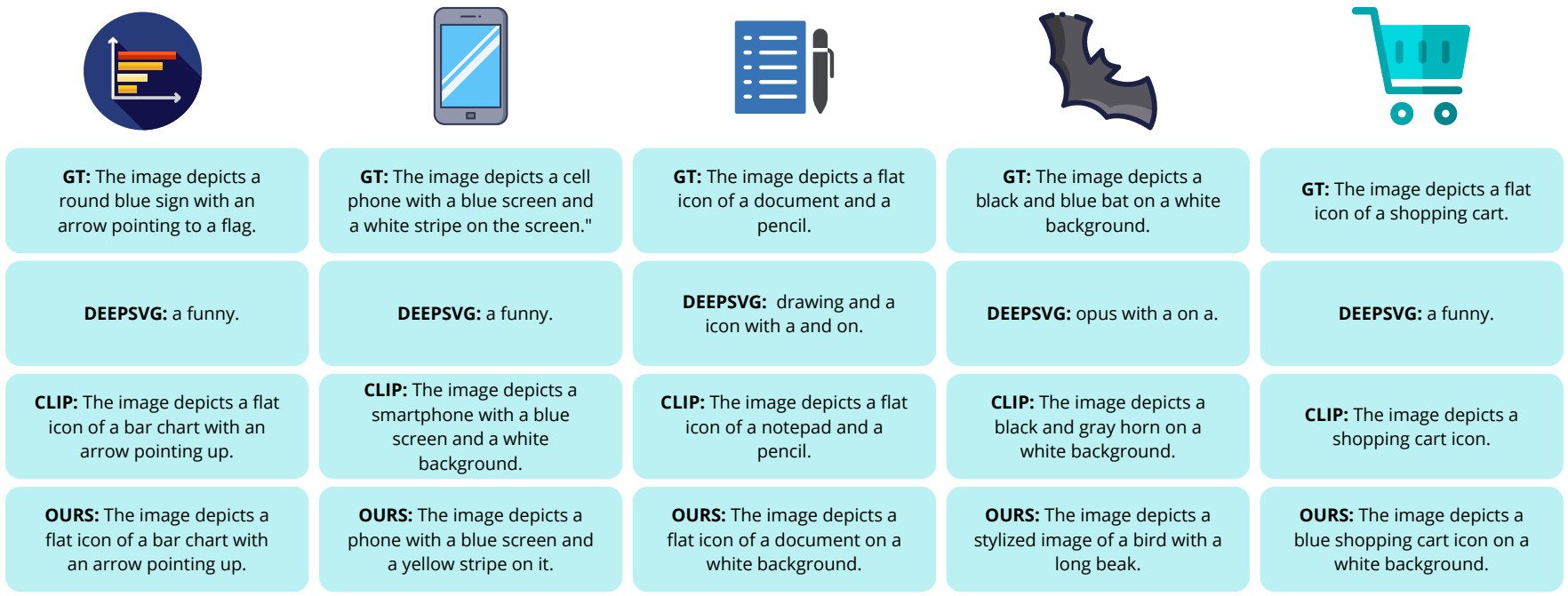

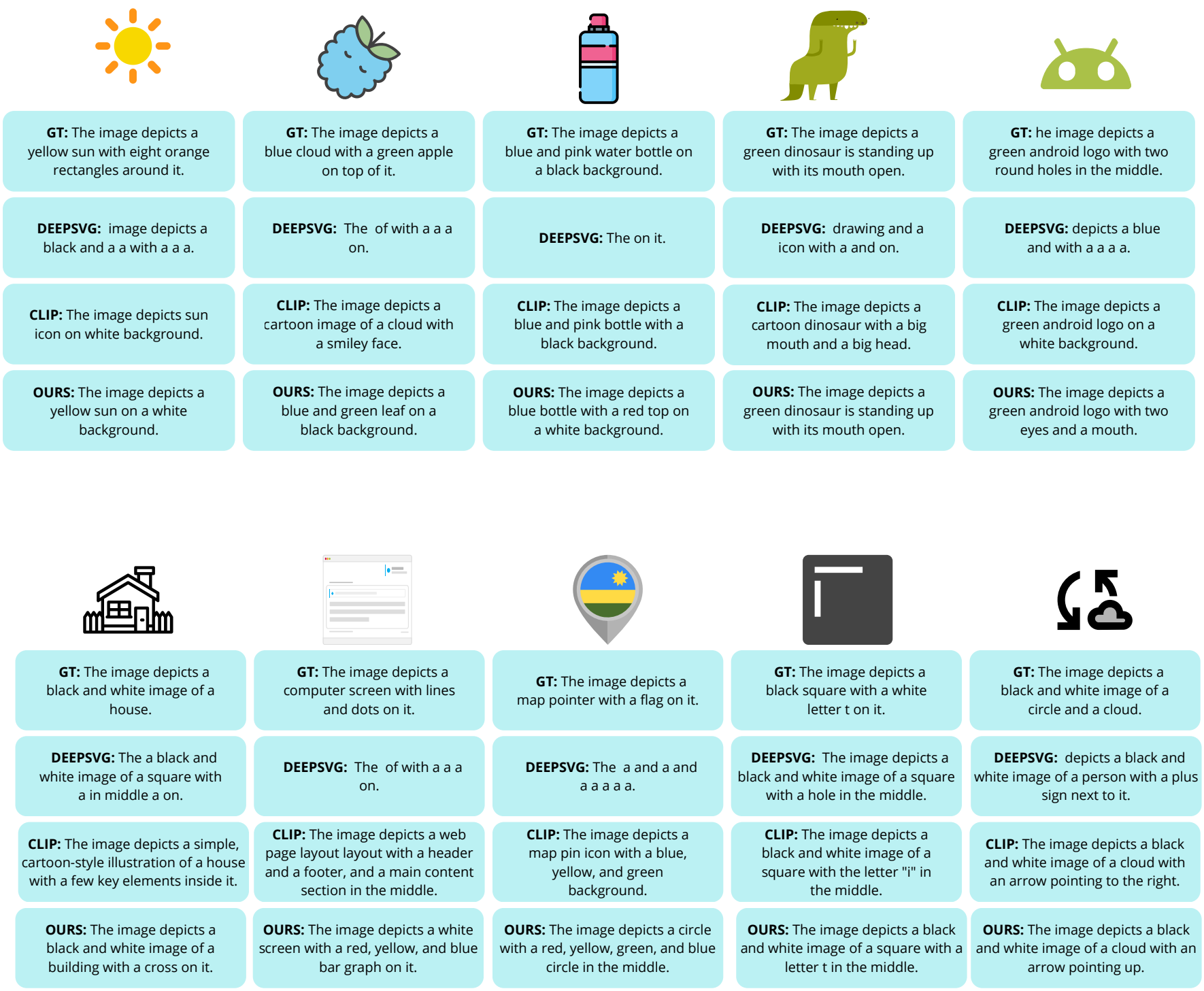


**Fig. 2:** Qualitative results on SVG captioning. We use Llama3.2 1B (first images row), Qwen3 0.6B (second images row) and Gemma2 2B (last images row) as language model and compare different image encoders: ground truth SVG (first row), DeepSVG, CLIP, and SLS (Ours). Each column represents a different example.

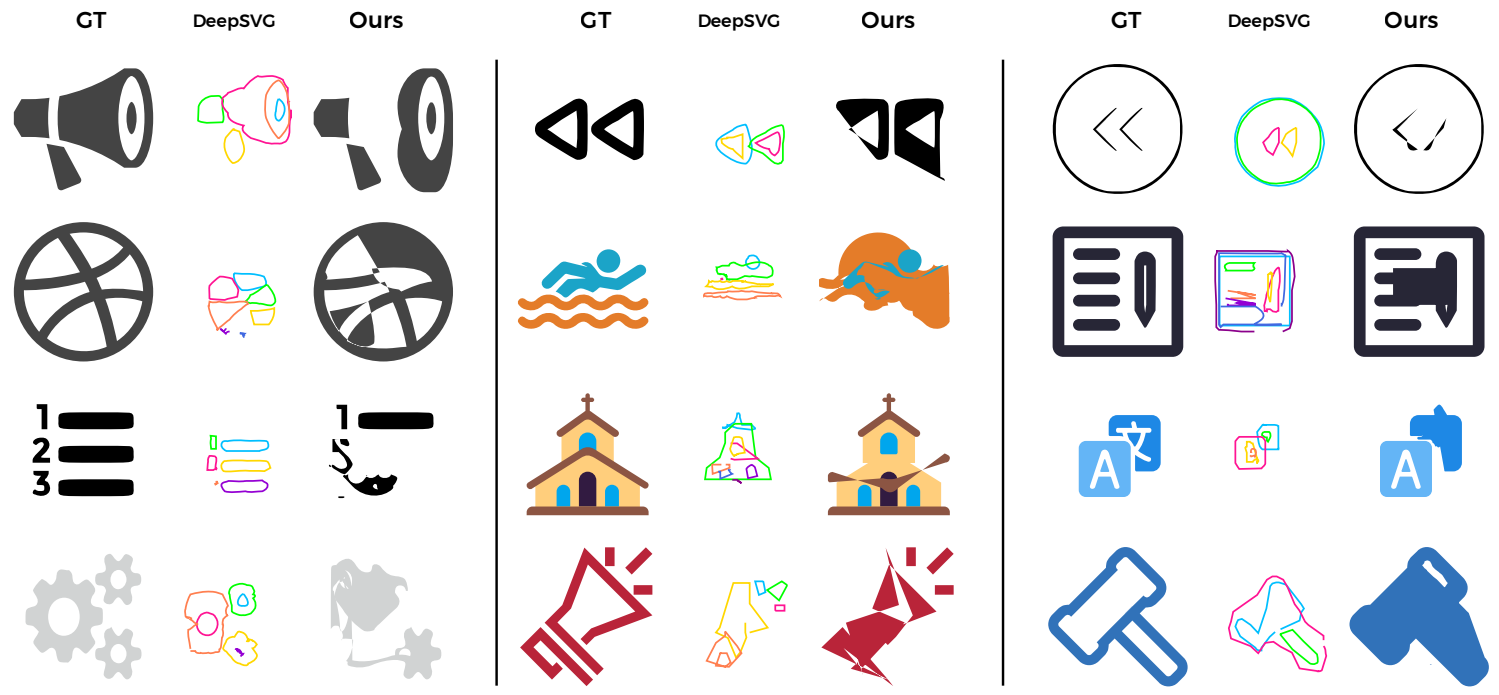


**Fig. 3:** Failure cases of SLS reconstruction. Each separated column shows three samples from left to right: ground truth (GT) SVG images, DeepSVG reconstructions, and SLS (Ours) reconstructions.

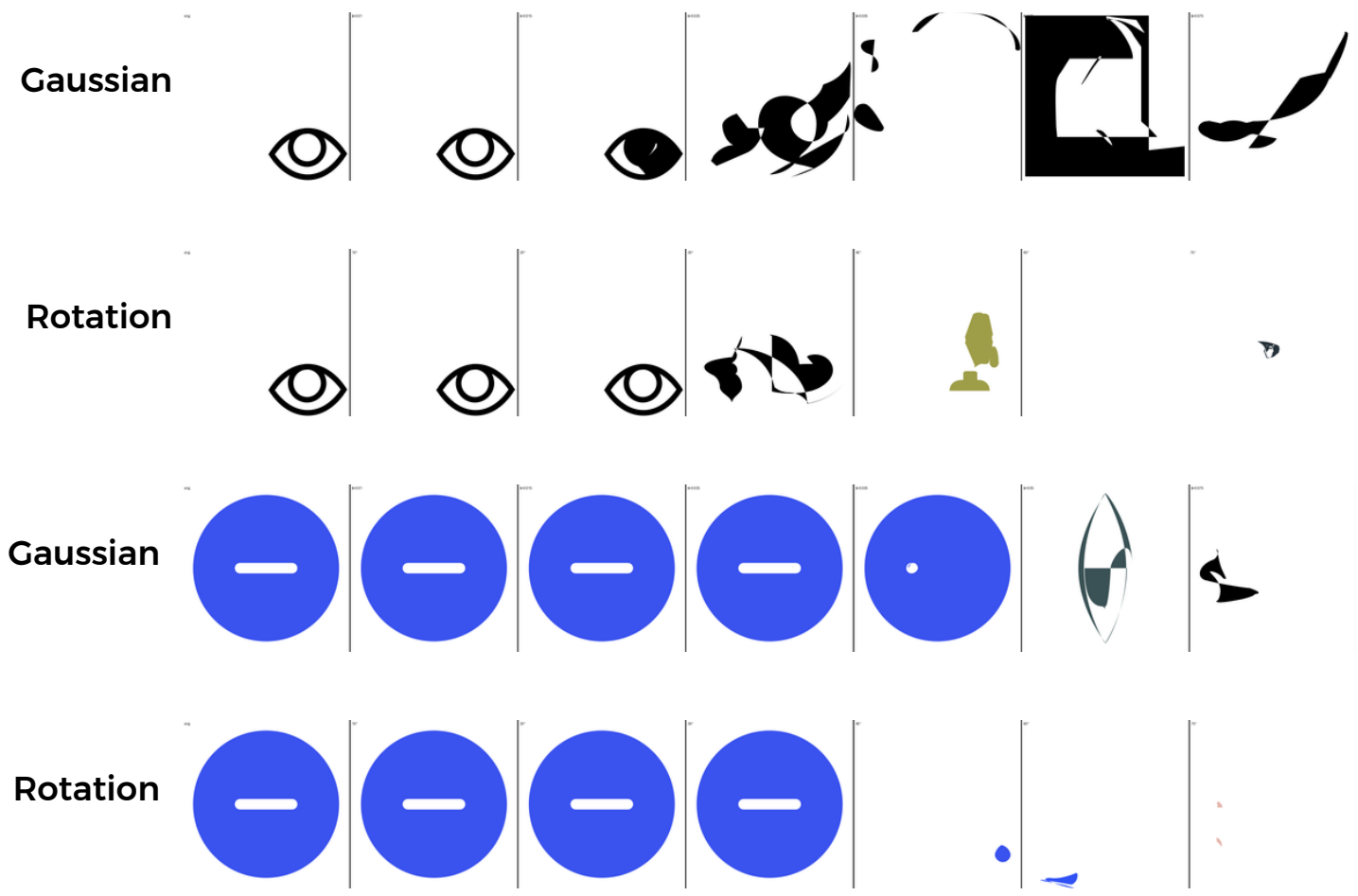


**Fig. 4:** Qualitative results on noise robustness. SVG paths subject to increasing rotation (up to 30°) and Gaussian noise ($\sigma = 0.02$). From left to right: original SVG, perturbed inputs with increasing gaussian noise magnitude and rotation angle. Results demonstrate robust geometric invariances within the learned representation, maintaining reconstruction fidelity under significant perturbations.

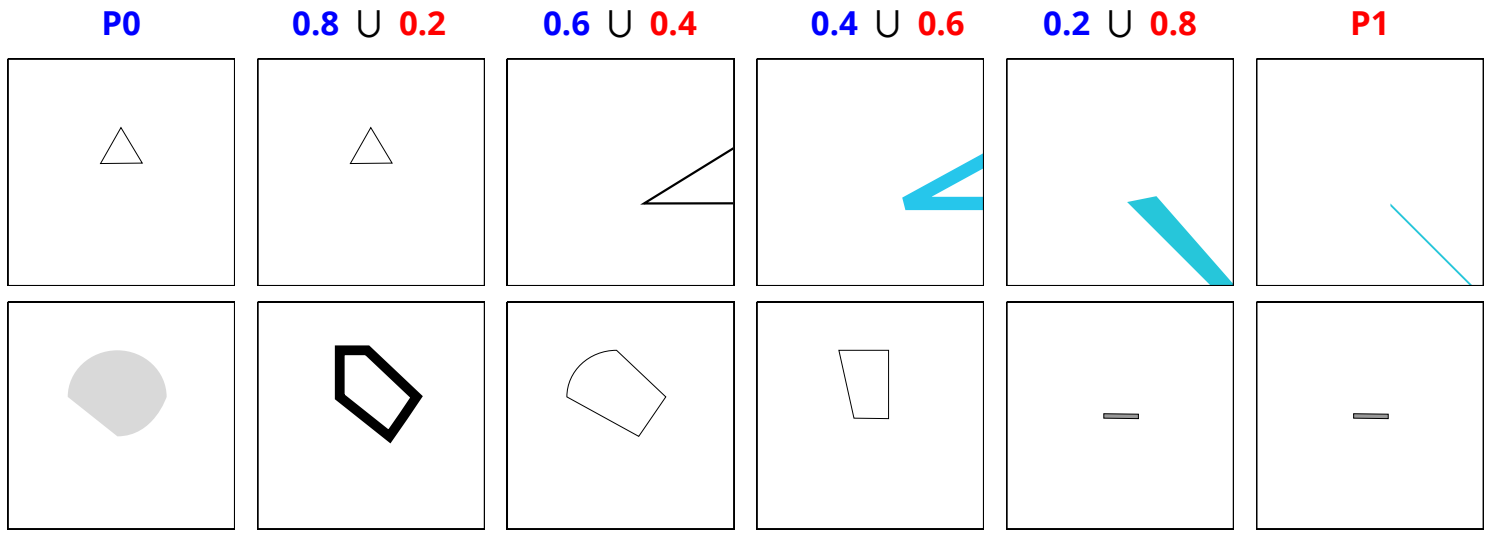


**Fig. 5:** Qualitative latent-space interpolation using spherical linear interpolation (SLERP). Each row interpolates between two SVG path embeddings ($P_0$ and $P_1$). The decoded paths evolve smoothly in geometry and style, illustrating the continuity of the learned latent space.

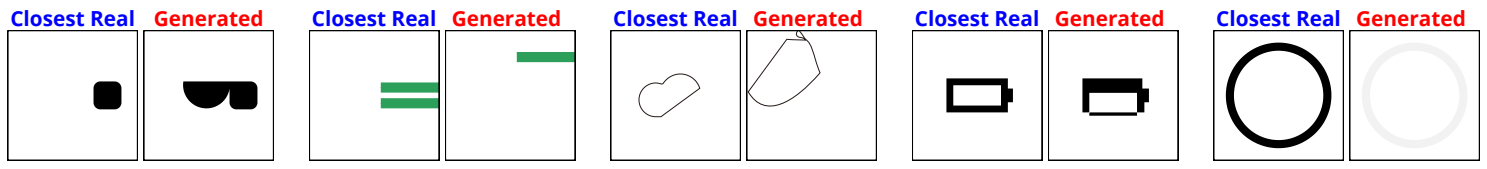


**Fig. 6:** Generation from latent-space neighborhoods. For each example, we randomly sample an embedding from the local neighborhood of a real path and decode it using the pretrained decoder. Generated paths remain valid and visually plausible while exhibiting meaningful variations with respect to their closest real counterparts.

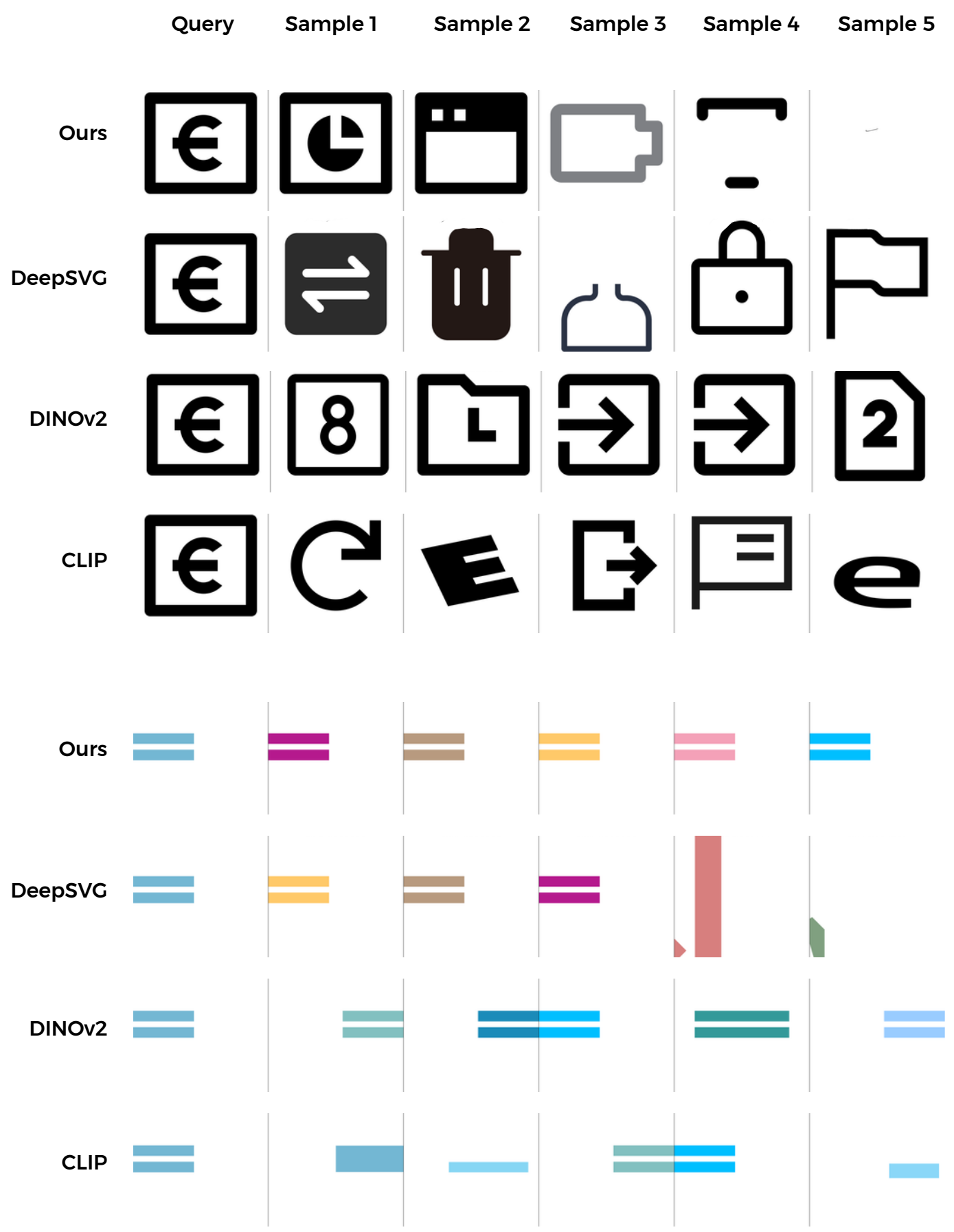


**Fig. 7:** Qualitative results on SVG retrieval. Query SVG path (left column) with top-5 ranked retrievals from our method compared to DeepSVG, DINOv2, CLIP baselines. Our method produces more semantically consistent results, demonstrating superior retrieval quality.


\end{document}